\documentclass[12pt]{article}
\usepackage{amsmath}
\usepackage{amssymb}
\usepackage{times}
\usepackage{graphicx}
\usepackage{color}
\usepackage{multirow}
\usepackage[authoryear]{natbib}
\usepackage{rotating}
\usepackage{bbm}
\usepackage{latexsym}
\usepackage{appendix}
\usepackage{cancel}

\usepackage{hyperref}
\usepackage{booktabs}
\usepackage{multirow}
\usepackage{algorithm}
\usepackage{algpseudocode}

\newcommand{\captionfonts}{\normalsize}

\makeatletter  
\long\def\@makecaption#1#2{%
  \vskip\abovecaptionskip
  \sbox\@tempboxa{{\captionfonts #1: #2}}%
  \ifdim \wd\@tempboxa >\hsize
    {\captionfonts #1: #2\par}
  \else
    \hbox to\hsize{\hfil\box\@tempboxa\hfil}%
  \fi
  \vskip\belowcaptionskip}
\makeatother   
\begin{document}
\hspace{13.9cm}

\ \vspace{20mm}\\

%{\LARGE Expectation Reflection Learning for Multilayer Neural Networks}
{\LARGE {Multiplicative learning from observation-prediction ratios}}

\ \\
{\bf \large Han Kim$^{\displaystyle 1}$, Hyungjoon Soh$^{\displaystyle 1}$, Vipul Periwal$^{\displaystyle 2}$, Junghyo Jo$^{\displaystyle 1, \displaystyle 3, \displaystyle 4, \displaystyle *}$}\\
{$^{\displaystyle 1}$Department of Physics Education, Seoul National University, Seoul 08826, Korea.}\\
{$^{\displaystyle 2}$Laboratory of Biological Modeling, National Institute of Diabetes and Digestive and Kidney Diseases, National Institutes of Health, Bethesda, Maryland 20892, USA.}\\
{$^{\displaystyle 3}$Center for Theoretical Physics and Artificial Intelligence Institute, Seoul National University, Seoul 08826, Korea.}\\
{$^{\displaystyle 4}$School of Computational Sciences, Korea Institute for Advanced Study, Seoul 02455, Korea.}\\
%{$^{\displaystyle 5}$The Abdus Salam International Center for Theoretical Physics, Strada Costiera 11, 34151 Trieste, Italy.}\\
{$^{\displaystyle *}$Corresponding author: jojunghyo@snu.ac.kr}\\
%

%\ \\[-2mm]

\thispagestyle{empty}
\newpage
\markboth{}{NC instructions}
%
%\\vspace{30mm}\\

%
%Abstract
\begin{center} {\bf Abstract} \end{center}
Additive parameter updates, as used in gradient descent and its adaptive extensions, underpin most modern machine-learning optimization. Yet, such additive schemes often demand numerous iterations and intricate learning-rate schedules to cope with scale and curvature of loss functions. Here we introduce Expectation Reflection (ER), a multiplicative learning paradigm that updates parameters based on the ratio of observed to predicted outputs, rather than their differences. ER eliminates the need for ad hoc loss functions or learning-rate tuning while maintaining internal consistency. Extending ER to multilayer networks, we demonstrate its efficacy in image classification, achieving optimal weight determination in a single iteration. We further show that ER can be interpreted as a modified gradient descent incorporating an inverse target-propagation mapping. Together, these results position ER as a fast and scalable alternative to conventional optimization methods for neural-network training.
%%%%%%%%%%

\section{Introduction}\label{sec1}
Artificial neural networks are built from perceptrons as basic computational units, inspired by the interconnected structure of the brain. In multilayer perceptrons, signals propagate through stacked layers, where each neuron aggregates weighted inputs from earlier nodes and applies a nonlinear transformation. These networks learn through error backpropagation (BP), an algorithm that adjusts connection weights by gradient descent to reduce the mismatch between predictions and observations~\citep{rumelhart1986learning}. BP effectively assigns credit across layers, enabling coordinated weight updates and stable learning in deep architectures. As the core training mechanism of modern neural systems, BP powers the large-scale AI models~\citep{brown2020language,ramesh2021zero,rombach2022high}.

Despite its effectiveness, the \emph{global} coordination required by BP is widely regarded as biologically implausible. In contrast, a common modeling assumption in neuroscience and biologically motivated learning is that synaptic updates are driven primarily by \emph{local} information available at each connection (e.g., presynaptic and postsynaptic activity), possibly modulated by a separate global context or reward signal~\citep{Hebb1949,Caporale2008,Fremaux2016}. To bridge this gap, researchers have explored alternative frameworks, ranging from dynamic recurrent models and predictive coding to target propagation and random feedback alignment, that seek to approximate BP while adhering more closely to biological constraints~\citep{pindeda1987,Crick1989,rao1999predictive,bengio2014auto,lillicrap2016random,Whittington2019,lillicrap2020backpropagation}. Among these, target propagation (TP) offers a promising direction by replacing explicit error transmission with locally defined target activations. Instead of relying on gradient signals, TP propagates approximate inverse mappings from later to earlier layers, enabling each neuron to update its weights using only local information. In doing so, TP retains much of the computational efficiency of BP while reducing its dependence on symmetric feedback pathways and global coordination~\citep{bengio2014auto}.

Even when its biological implausibility is set aside, BP is limited in computational efficiency. In convex optimization, methods that incorporate information about scale and curvature of loss functions, such as natural gradients~\citep{Amari1998} and mirror descent~\citep{NemirovskiYudin1983, BeckTeboulle2003, Soh2023MirrorHopfield}, have been developed to accelerate convergence and improve stability. Nevertheless, the \emph{additive} nature of BP remains the default optimization scheme in modern machine learning. Additive steps are sensitive to scaling and curvature, often requiring elaborate learning-rate schedules and normalization heuristics to achieve reliable convergence in deep or ill-conditioned settings. By contrast, \emph{multiplicative} learning rules are inherently scale-aware and positivity-preserving: they update parameter ratios rather than differences. This principle underlies classical algorithms such as the exponentiated gradient for online prediction~\cite{KivinenWarmuth1997}, multiplicative updates for nonnegative matrix factorization~\cite{LeeSeung2001}, and generalized iterative scaling for log-linear models~\cite{DarrochRatcliff1972}.

In the same spirit as these multiplicative approaches, we developed the Expectation Reflection (ER) method as a novel framework for regression-based learning~\citep{hoang2019network}. Unlike conventional regression, which adjusts model parameters based on the difference between predicted and observed outputs, ER uses their ratio as the basis for learning. This ratio-based update offers a natural mechanism for adjusting the scale of model parameters: when predictions exceed observations, the parameters are reduced, and when predictions fall short, they are amplified. The same principle extends to the sign of the relationship, allowing parameters to reverse direction when the prediction and observation differ in polarity. By relying on ratios rather than differences, ER embodies a form of multiplicative learning that is inherently adaptive to scale, offering something of a more efficient and stable alternative to traditional additive updates.
However, ER is currently limited to regression problems and applies only to neural networks with input and output layers, without intermediate hidden layers.

In this work, we extend ER to multilayer neural networks. We derive a multilayer ER learning rule and show that it can be interpreted as a multiplicative analogue of BP with modified update dynamics. This formulation preserves the efficiency of ER while enabling learning in deep architectures. Through numerical experiments, we demonstrate that the resulting algorithm achieves faster convergence and competitive performance compared with conventional gradient-based training, highlighting ER as a scalable alternative to standard backpropagation for multilayer learning.

%Our goal is to extend this efficient learning method to a multilayer learning algorithm. In this study, we formulate the multilayer ER algorithm, and show that ER is closely related to BP with some modifications that enable faster learning. Finally, we present experimental results that showcase the efficiency of ER in multilayer configurations.This demonstrates ER as a fast and scalable alternative to conventional gradient descent methods.

%This contrasts with the prevailing view that ER is a completely distinct alternative to BP. Furthermore, we highlight its connection to TP, providing deeper insight into its underlying mechanisms and establishing a foundation for its seamless extension to multilayer settings. Notably, this connection indicates that ER follows a \emph{local} learning rule while achieving high efficiency due to its \emph{multiplicative} parameter update. Finally, we present experimental results that showcase the efficiency of ER in multilayer configurations.

\section{Related works}
To clarify the concept of multiplicative learning based on observation-prediction ratios rather than differences, we briefly introduce prior work on backpropagation, target propagation, and expectation reflection, using consistent notation.

\subsection{Backpropagation}
Consider a series of transformations that map an input signal $X$ to an output signal $Y$, represented as:
\begin{equation*}
    X=Z_0 \to Z_1 \to Z_2 \to \cdots \to Z_{L-1} \to Z_L = Y,
\end{equation*}
where $Z_0$ and $Z_L$ denote the input and output, respectively, for notational convenience.
Signal propagation consists of two key components:
(i) the pre-activation $S_l$ at layer $l$, computed as the weighted sum of presynapctic activity $Z_{l-1}$, and (ii) its nonlinear transformation:
\begin{align}
	S_l &= Z_{l-1}W_l,\nonumber\\
	Z_{l} &= \sigma(S_l).
\label{eq:MLNN}
\end{align}
Here, $\sigma(S_l)$ represents a nonlinear activation function applied to the pre-activation $S_l$.
The sequential transformation ultimately produces the output $Z_L=Y$. We then define the error between the predicted output $Y$ and the observed output $\hat{Y}$. This discrepancy is quantified by a loss function, $\mathcal{L}(\hat{Y}, Y)$.
Given the loss function, the weight parameters can be iteratively updated in the direction of the steepest descent using the gradient descent method:
\begin{equation}\label{eq:BPUpdate}
    W^{'}_l = W_l - \eta \frac{\partial \mathcal{L}}{\partial W_l},
\end{equation}
where $\eta$ is the learning rate, and $l$ denotes the layer index, ranging from $1$ to $L$.

BP offers an efficient method to compute gradients of hidden layers~\citep{rumelhart1986learning}. Using the chain rule, the negative gradient is given by:
\begin{align*}
     -\frac{\partial \mathcal{L}}{\partial W_l} =-\frac{\partial  Z_l}{\partial W_l}\frac{\partial  Z_{l+1}}{\partial Z_l}\cdots\frac{\partial Z_L}{\partial Z_{L-1}}\frac{\partial \mathcal{L}}{\partial Z_L}.
\end{align*}
The changes in activities and pre-activations are computed iteratively as follows:
\begin{subequations}\label{eq:BPonMNN}
    \begin{align}
        \Delta Z_L &= -\frac{\partial \mathcal{L}}{\partial Z_L}, \label{eq:BPonMNN1}\\
        \Delta S_L &=  \frac{\partial Z_L}{\partial S_L}\odot\Delta Z_L,\label{eq:BPonMNN2}\\
        \Delta Z_{l} &= \Delta S_{l+1} W_{l+1}^T,\label{eq:BPonMNN3}\\
        \Delta S_{l} &=  \frac{\partial Z_{l}}{\partial S_{l}}\odot\Delta Z_{l},\label{eq:BPonMNN4}
    \end{align}
\end{subequations}
where $l$ iterates from $L-1$ to $1$, and $\odot$ represents the element-wise Hadamard product.
%Here, $\sigma'(S_l) \equiv \partial Z_l / \partial S_l$ denotes the gradient of the activation function, and $\odot$ represents the element-wise Hadamard product.
Then, the weight update for the $l$-th layer is formulated as
\begin{equation}
     -\frac{\partial \mathcal{L}}{\partial W_l} = Z_{l-1}^T\Delta S_l,
    \label{eq:BP}
\end{equation}
which is known as the delta rule.
%Notably, the weight update at layer $l$ depends only on local signals, the neighboring pre-activation difference $\Delta S_l$ and the presynaptic activity $Z_{l-1}$. When the output error $\Delta Z_L$ propagates backward through the network, the weight update appears to follow a local learning rule.

\subsection{Target Propagation}\label{subsec:TP}
TP was proposed as an alternative credit-assignment scheme that avoids the strict ``weight-transport'' requirement of standard BP, namely that feedback pathways effectively implement the transpose of forward weights in Eq.~\eqref{eq:BPonMNN3}. Instead of using transposed weights, TP propagates layerwise targets through approximate inverses of the forward mappings~\citep{bengio2014auto,lee2015difference,meulemans2020theoretical}. 

Given a forward mapping $f_l$, TP learns an approximate inverse mapping $g_l$ such that
\begin{align*}
    Z_l = f_l(Z_{l-1}; W_l), \phantom{MM}
    Z_{l-1} \approx g_l(Z_l; V_l),
\end{align*}
where $V_l$ is trained by minimizing an inverse (reconstruction) loss that encourages $g_l(Z_l; V_l)$ to reconstruct $Z_{l-1}$.
Once these inverse models are learned, TP defines layerwise targets by taking a small step at the output (e.g., $T_L = Y - \eta \partial \mathcal{L}/ \partial Y$) and then mapping these targets backward through the learned inverses,
\begin{align}
    T_{l-1} &= g_l(T_l; V_l) \phantom{MM} (l=L, \ldots, 2)\label{eq:TPBack},
\end{align} 
followed by local layer updates that reduce a local discrepancy $\mathcal{L}_l(Z_l, T_l)$ between the current activity $Z_l$ and its target $T_l$~\citep{bengio2014auto,lee2015difference}.

A key limitation of pure TP is that it relies on the inverse networks being sufficiently accurate. When $g_l$ only approximates $f_l^{-1}$, the mismatch (``inverse error'') can accumulate across layers.
As a result, pure TP is generally not guaranteed to produce updates aligned with the gradient of the global loss, nor to converge to a local minimum of $\mathcal{L}$ in typical deep-network settings~\citep{lee2015difference,meulemans2020theoretical}. 

Difference Target Propagation (DTP)~\citep{lee2015difference} addresses this inverse-error issue by propagating differences through the same approximate inverse. The core observation is that even if $T_l = Z_l$, it does not necessarily follow that $T_{l-1} = Z_{l-1}$, because $g_l$ is only an approximation of $f_l^{-1}$ in $T_{l-1} = g_l(T_l; V_l)$ whereas $Z_{l-1} = f_l^{-1}(Z_l; W_l)$.
Therefore, DTP constructs targets so that the propagated difference satisfies
\begin{align*}
	T_{l-1} - Z_{l-1} = g_l(T_l; V_l) - g_l(Z_l; V_l),
 \end{align*}
which leads to the corrected target
\begin{align}
	T_{l-1}& = g_l(T_l; V_l) - \underbrace{\Big(g_l(Z_{l};V_l) - Z_{l-1}\Big)}_\text{error adjustment term}.
	%&= g_l(T_l; V_l) - \Big(g_l\big(f_l(Z_{l-1};W_l);V_l\big) - Z_{l-1}\Big)\quad(l= L,\cdots,1).
\end{align}
By construction, this correction enforces $T_{l-1} = Z_{l-1}$ when $T_l = Z_l$ (up to inverse approximation quality), and empirically improves stability over pure TP~\citep{lee2015difference}.
Even with the difference correction, DTP does not generally recover exact BP gradients in arbitrary deep networks, and establishing broad convergence guarantees remains an open problem~\citep{meulemans2020theoretical}.

\subsection{Expectation Reflection}
Hoang \textit{et al.} introduced ER~\citep{hoang2019network}.
Though it originated from spin models in statistical physics, ER can be universally applied to learn the logistic regression relationship between input and output in any dataset. Given the data, $\{X, \hat{Y}\}$, the ER algorithm comprises forward and backward processes. The forward process is represented as:
\begin{align}
	S &= XW, \nonumber \\
	Y &= \sigma(S) = \tanh(S).
\end{align} 
Here, the hyperbolic tangent function is used as the activation function $\sigma(S)$ to represent binary outputs.
The backward process is defined to update the pre-activations:
\begin{equation}
    S^{'} = \frac{\hat{Y}}{Y}\odot S,\label{eq:ER_update}
\end{equation}
where the division is performed element-wise. 
This update is designed to align the predicted output $Y$ and the observed output $\hat{Y}$. 
When $\hat{Y}$ exceeds or falls below $Y$, the corresponding pre-activation increases or decreases. If $\hat{Y}$ and $Y$ have opposite signs, the pre-activation flips accordingly.
This adjustment projects the observed data $\hat{Y}$ onto the model’s expectation $Y$, which is the principle underlying the name ``Expectation Reflection.''

Another notable aspect is the unique property of the hyperbolic tangent function. As the pre-activation $S$ approaches zero, the expected output $Y=\tanh(S)$ also approaches zero. This leads to a singular condition in the update equation due to division by zero. However, the update naturally avoids this issue thanks to the limiting behaviour:
\begin{equation*}
    \lim_{S\rightarrow0} \frac{\hat{Y}}{Y} \odot S = \lim_{S\rightarrow0} \hat{Y} \odot \frac{S}{Y} = \lim_{S\rightarrow0} \hat{Y} \odot \frac{S}{\tanh(S)} = \hat{Y}.
\end{equation*}
This property ensures numerical stability, allowing smooth and reliable updates even in near-zero conditions. One can also obtain this property by using the error function albeit with a constant rescaling~\citep{Han2021}.

Notably, ER employs multiplicative (ratio-based) corrections rather than purely additive gradient steps, which can accelerate learning in practice. Moreover, ER can be implemented without the learning-rate schedules and explicit loss design typically required by gradient-descent training, leading to a simpler and more self-contained optimization procedure.

Finally, after updating $S$ to an improved value $S^{'}$, the weight parameters are adjusted using linear regression in $S^{'}=X W^{'}$:
\begin{equation}
    W^{'} = X^\dagger S^{'},\label{eq:SingleLayerERBack}
\end{equation}
where $X^\dagger = (X^T X)^{-1}X^T$ is the Moore-Penrose pseudo-inverse of $X$~\citep{Penrose1955}.

\section{Theory}
We extend ER to multilayer networks and adapt the original formulation using the idea of TP.
Finally, we express the framework in terms of BP.
This formulation highlights the key differences between BP and ER.

\begin{figure}[h]
\centering
\includegraphics[width=\textwidth]{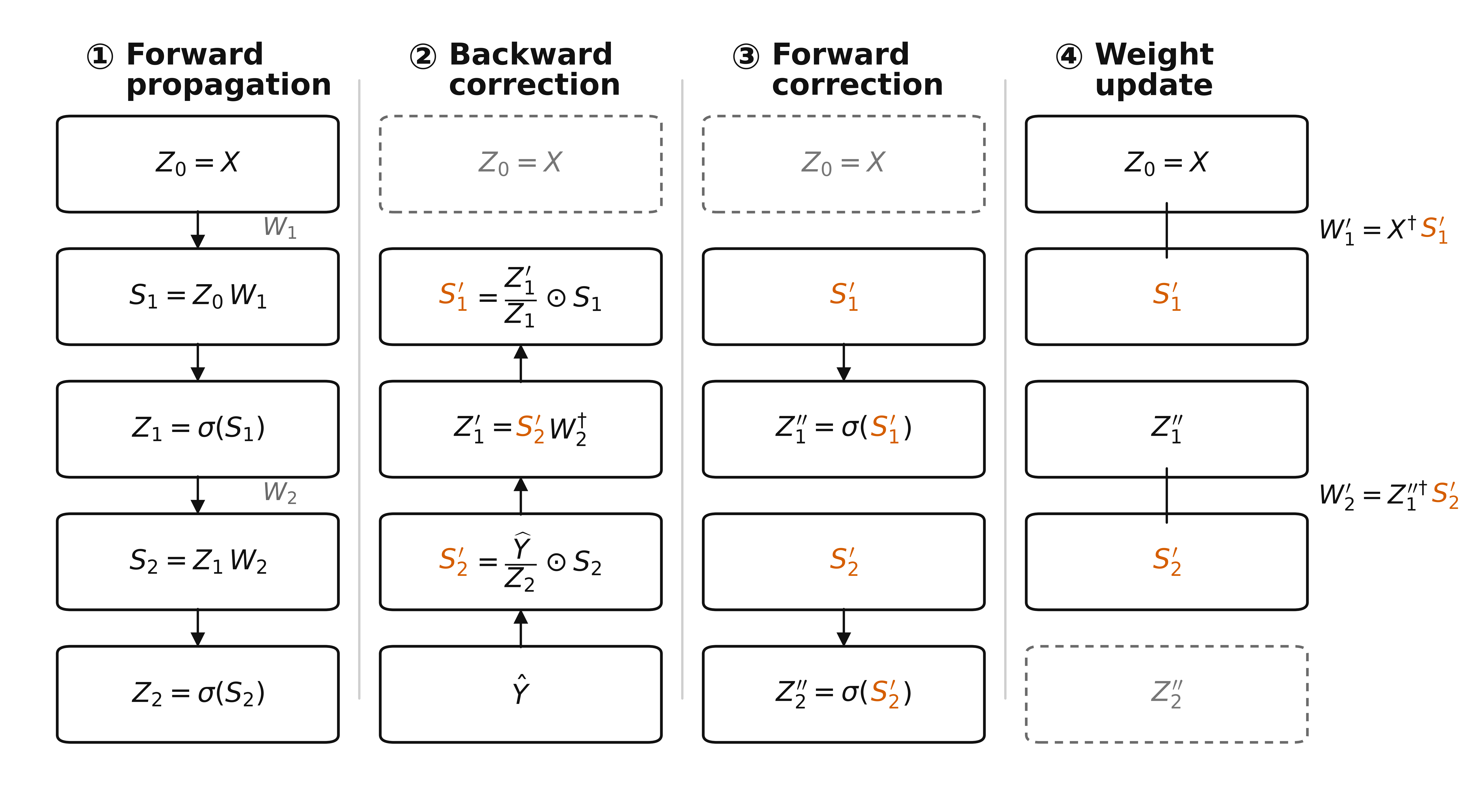}
\caption{Multilayer Expectation Reflection. Starting form the input $Z_0 = X$, signals propagate forward through the hiden layer (layer 1) to the output layer (layer 2) (Forward propagation). Given the target $\hat{Y}$, the pre-activities $S_l$ and post-activities $Z_l$ are corrected backward (Backward correction). The post-activities are then recomputed from the corrected pre-activities (Forward correction). Finally, weights are updated locally using the corrected neighboring post- and pre-activities (Weight update). This schematic shows a single hidden layer for clarity, but the procedure extends directly to networks with an arbitrary number of intermediate layers.}\label{fig1}
\end{figure}

\subsection{Multilayer expansion}
The forward process in multilayer ER proceeds straightforwardly, analogous to that in conventional deep neural networks (see Eq.~(\ref{eq:MLNN})).
In contrast, the backward process presents challenges, as single-layer ER does not involve error propagation.
When the forward process is not fully trained, the predicted output $Z_L$ may deviate from the observed output $\hat{Y}$.
To reduce this discrepancy, the pre-activation $S_L$ is refined using the ratio-based multiplicative update rule of ER:
\begin{align}
    S^{'}_L = \frac{\hat{Y}}{Z_L} \odot S_L. 
\end{align}
This refinement propagates backward by updating $Z_{L-1}$ via the relation ${S}_L = Z_{L-1} W_L$:
\begin{align}
\label{eq:Z_update}
    Z^{'}_{L-1} = S^{'}_L W_L^\dagger, 
\end{align}
where $W_L$ remains fixed during this step.
The updated $Z^{'}_{L-1}$ then enables correction of the preceding pre-activation:
\begin{align}
    S^{'}_{L-1} = \frac{Z^{'}_{L-1}}{Z_{L-1}} \odot S_{L-1}. 
\end{align}
Repeating this process layer by layer ultimately updates $S_1$, completing the backward refinement.
The post-activity $Z^{'}_l$ is then recalculated from the refined preactivation $S^{'}_l$ as
\begin{align}
    Z^{''}_{l} = \sigma(S^{'}_{l}), 
\end{align}
where $l=1, \ldots, L-1$.

Finally, using the refined pre-activations and corresponding post-activities, the weights are updated locally as
\begin{align}
\label{eq:W_update}
    W^{'}_{l} = Z^{''\dagger}_{l-1} S^{'}_l, 
\end{align}
with $l = 1, \ldots, L$.
This update rule for $W_l$ is strictly \emph{local}, relying only on the neighboring variables $Z^{''}_{l-1}$ and $S^{'}_l$.
It is important to note that since all updates stem from the refined $S^{'}_L$, the resulting output $Z^{''}_L$ progressively converges toward the target $\hat{Y}$.

Recalculation of $Z^{''}_{l-1}$ is essential for two reasons.
First, if $Z^{'}_{l-1}$ were used without recalculation, the weight update becomes ineffective, as $S^{'}_l = Z^{'}_{l-1} W_l$ would yield no change in $W_l$.
Second, this recalculation mitigates {\it internal covariate shift}, the phenomenon where updates in lower-layer weights alter the input to higher layers, degrading performance when static input assumptions are made~\citep{ioffe2015batch}.
By updating each layer's weights only after revising the input activities of preceding layers, the method naturally accommodates changes in input across the network.

Practically, however, this algorithm exhibits poor performance and unstable training.
The degradation likely arises from inaccuracies in computing the pseudo-inverse, which introduce regression errors.
Consequently, even if $S^{'}_{l+1} = S_{l+1}$, the equality $Z^{'}_{l} = Z_{l}$ is not guaranteed. Such errors accumulate across layers, leading to substantial deviations in the output.
Reducing this inverse error is therefore crucial for stable and effective training.
Inspired by the difference target propagation (DTP) framework~\citep{lee2015difference}, we introduce an inverse-error adjustment term:
\begin{align}
	Z^{'}_{l} &= S^{'}_{l+1} W_{l+1}^\dagger- (S_{l+1} W_{l+1}^\dagger-Z_{l}),
    \label{eq:z_adjustment}
\end{align}
which replaces Eq.~(\ref{eq:Z_update}).
Similarly, to correct inverse errors during weight updates, we apply the adjustment:
\begin{align}
     	W^{'}_l &= Z_{l-1}^{''\dagger} S^{'}_l - (Z_{l-1}^{''\dagger} S_l-W_l),
        \label{eq:w_adjustment}
\end{align}
which replaces Eq.~(\ref{eq:W_update}).
Together, these adjustments complete the extension of ER to the multilayer setting.

\subsection{Reformulating Multilayer ER}
We now reformulate the multilayer ER in a manner analogous to BP, making it explicit how ER accelerates learning compared to BP.

\subsubsection{Comparison with BP}
First, we express the backward correction as the difference between the post- and pre-correction values:
\begin{subequations} \label{eq:diff}
\begin{align}
	 \Delta Z_L &= Z_L^{'} -  Z_L = \hat{Y} - Y \label{eq:diff1}\\
	 \Delta S_L &= S_L^{'} - S_L = \frac{Z_L^{'}}{Z_L} \odot S_L - \frac{Z_L}{Z_L} \odot S_L = \frac{S_L}{Z_L} \odot \Delta Z_L \label{eq:diff2}\\
	\Delta Z_{l} &= Z_{l}^{'} - Z_{l} = S_{l+1}^{'} W_{l+1}^\dagger - S_{l+1} W_{l+1}^\dagger = \Delta S_{l+1} W_{l+1}^\dagger \label{eq:diff3}\\
	\Delta S_{l} &= S_{l}^{'} - S_{l} = \frac{Z_{l}^{'}}{Z_{l}} \odot S_{l} - \frac{Z_{l}}{Z_{l}} \odot S_{l} = \frac{S_{l}}{Z_{l}} \odot \Delta Z_{l}, \label{eq:diff4}
\end{align}
\end{subequations}
where $l$ iterates from $L-1$ to $1$.
It is noteworthy that Eq.~\eqref{eq:diff3} coincides with the post-activity update incorporating the inverse-error adjustment in Eq.~\eqref{eq:z_adjustment}.

Second, the weight update with the inverse-error adjustment in Eq.~\eqref{eq:w_adjustment} can be rewritten in terms of the difference
\begin{align}
    \Delta W_l = W_l^{'} - W_l = Z_{l-1}^{''\dagger} S_l^{'} - Z_{l-1}^{''\dagger} S_l = Z_{l-1}^{''\dagger} \Delta S_l.
\end{align}
Accordingly, the weight update takes the additive form
\begin{equation*}
W_l^{'} = W_l + \Delta W_l.
\end{equation*}

We are now ready to make a direct, side-by-side comparison between BP and ER.
Their respective backward correction and weight update equations are summarized as follows:

\begin{subequations}
    \begin{minipage}{.5\textwidth}
        \begin{align*}
             & \text{BP} \nonumber \\
            \Delta Z_L &= -\frac{\partial \mathcal{L}(\hat{Y}, Z_L)}{\partial Z_L}\\
            \Delta S_L &=  \frac{\partial Z_L}{\partial S_L}\odot\Delta Z_L\\
            \Delta Z_{l} &= \Delta S_{l+1} W_{l+1}^T \\
            \Delta S_{l} &= \frac{\partial Z_{l}}{\partial S_{l}}\odot\Delta Z_{l}\\
            \Delta W_l &=  \eta Z_{l-1}^T \Delta S_l,
        \end{align*}
    \end{minipage}
    \begin{minipage}{.45\textwidth}
        \begin{align}
            & \text{ER} \nonumber \\
            \Delta Z_L &= \hat{Y}-Z_L\label{eq:ERBack1}\\
            \Delta S_L &= \frac{S_L}{Z_L}\odot\Delta Z_L\label{eq:ERBack2}\\
            \Delta Z_{l} &= \Delta S_{l+1} W_{l+1}^\dagger  \label{eq:ERBack3}\\
            \Delta S_{l} &= \frac{S_{l}}{Z_{l}}\odot\Delta Z_{l} \label{eq:ERBack4}\\
            \Delta W_l &= Z_{l-1}^{''\dagger} \Delta S_l. \label{eq:ERBack5}
        \end{align}
    \end{minipage} \vspace{\belowdisplayskip}
\end{subequations}

Here, in the backward correction Eqs.~\eqref{eq:ERBack3} and \eqref{eq:ERBack4}, the layer index runs from $l = L-1$ down to $l = 1$, while in the weight-update Eq.~\eqref{eq:ERBack5}, $l$ runs from $1$ to $L$.
By comparing these two sets of equations, we can interpret ER as a modified version of BP with the following differences:
\begin{enumerate}
	\item[(i)] The gradient of the loss function is replaced by the difference between the observed and the predicted output in Eq.~\eqref{eq:ERBack1};
	\item[(ii)] The slope $\partial Z_{l}/\partial S_l$ is replaced by $S_l/Z_l$ in Eqs.~\eqref{eq:ERBack2} and \eqref{eq:ERBack4};
	\item[(iii)] The transpose $^T$ is replaced by the pseudo-inverse $^\dagger$ in Eqs.~\eqref{eq:ERBack3} and \eqref{eq:ERBack5};
    \item[(iv)] The learning rate $\eta$ is fixed as $\eta = 1$ in Eq.~\eqref{eq:ERBack5};
    \item[(v)] The post-activity $Z_{l-1}$ is replaced by the updated post-activity $Z_{l-1}^{''}$ in Eq.~\eqref{eq:ERBack5}.
\end{enumerate}
Point (i) is noteworthy: unlike BP, ER does not require an {\it ad hoc} choice of loss function for the output error. It relies solely on the discrepancy (or consistency) between the observed output $\hat{Y}$ and the predicted output $Z_L$, which is equivalent to adopting the $\ell_2$ loss $\mathcal{L}(\hat{Y}, Z_L) = \tfrac12||\hat{Y}-Z_L||_2^2$.
The remaining modifications, points (ii), (iii), (iv), and (v), contribute to accelerating the learning process.
In particular, point (iv) directly speeds up learning, since BP typically employs small learning rates ($\eta \ll 1$).
In what follows, we explain in detail how points (ii), (iii), and (v) further contribute to accelerating learning.

\subsubsection{Scaling property of TP}
We note that if $S_{l}$ and $Z_{l}$ carry physical dimensions, the pre-activation update in BP fails to maintain dimensional consistency between the left-hand and right-hand sides in Eq.~\eqref{eq:ERBack4}.
In contrast, the pre-activation update in ER maintains the dimensional consistency in Eq.~\eqref{eq:ERBack4}.
Consider that $Z_{l}=\sigma(S_{l})$ is a monotonically increasing function passing through the origin at $(0, 0)$.
The ratio
\begin{equation*}
    \frac{Z_{l}-0}{S_{l}-0} \approx \frac{\partial Z_{l}}{\partial S_{l}}
\end{equation*}
serves as a na\"ive approximation of the instantaneous slope at $S_{l}$.
This leads to a novel interpretation of the pre-activation updates in Eq.~\eqref{eq:ERBack4}:
\begin{align*}
    \Delta S_{l} = \frac{S_{l}}{Z_{l}} \odot \Delta Z_{l} \approx \bigg[ \frac{\partial Z_{l}}{\partial S_{l}} \bigg]^{-1} \odot \Delta Z_{l}. \\
\end{align*}
The use of the inverse activation function is reminiscent of TP. We now demonstrate an explicit link between ER and TP. 
We assume that the inverse function $g_l = f_l^{-1}$ represents the exact inverse of the activation function $f_l$:
\begin{align*}
    Z_l &= f_l(Z_{l-1}; W_l) = \sigma(Z_{l-1} W_l), \\
    Z_{l-1} &= g_l(Z_l; V_l)=\sigma^{-1}(Z_l) W_l^{-1}.
\end{align*}
Once the target at layer $l$ has been updated to $T_l = Z_l^{'}$, we examine how this target signal propagates to the lower layer, yielding the activation target $T_{l-1} = Z_{l-1}^{'}$.
For a small deviation $\Delta Z_l = Z_l' - Z_l$, a first–order Taylor expansion of $g_l$ around $Z_l$ gives
\begin{align*}
    Z_{l-1}^{'} &= g_l(Z_l^{'}; V_l) \\
    &= g_l(Z_l; V_l) + \frac{\partial g_{l}}{\partial Z_l} \odot \Delta Z_l \\
    &= Z_{l-1} + \frac{\partial \sigma^{-1}(Z_l)}{\partial Z_l} \odot \Delta Z_l W_l^{-1}.
\end{align*}
Using $S_l = \sigma^{-1}(Z_l)$, this can be rewritten as
\begin{align*}
    (Z_{l-1}^{'} - Z_{l-1}) W_l &=  \frac{\partial S_l}{\partial Z_l} \odot \Delta Z_l \\
    S_l^{'} - S_l &= \Delta S_{l} = \bigg[ \frac{\partial Z_{l}}{\partial S_{l}} \bigg]^{-1} \odot \Delta Z_l.
\end{align*}
This expression coincides with the pre-activation update used in ER.
Thus, ER can be viewed as implementing a linearized target-propagation step based on an (approximate) inverse mapping, making the link between ER and TP explicit.

Beyond the scaling property of TP, the pre-activation update in ER, employing $S_l/Z_l$ instead of $\partial Z_l/\partial S_l$ in Eq.~\eqref{eq:ERBack4}, offers a practical advantage for error-signal propagation.
In our setting, the layer output is $Z_l= \sigma(S_l) =\tanh(S_l)$.
Since $S_l/Z_l=S_l/\tanh(S_l) \ge 1$ for all $S_l$, while $\partial Z_l/\partial S_l = \textrm{sech}^2(S_l) \le 1$, the ER pre-activation update amplifies the error signal $\Delta Z_l$, whereas the BP update attenuates $\Delta Z_l$, potentially slowing down or hindering learning.
Moreover, when the network predicts an output with the wrong sign, the derivative $\partial Z_l/\partial S_l$ loses the error signal, while $S_l/Z_l$ retains it. For instance, suppose the target $Z_l^{'} = 1$ but the predicted output $Z_l = -1$ with $S_l \rightarrow -\infty$. In this scenario, $\partial Z_l/\partial S_l = \textrm{sech}^2(S_l) \rightarrow 0$, leading to $\Delta S_l \rightarrow 0$ and impeding learning. By contrast, using $S_l/Z_l$ yields $\Delta S_l \rightarrow \infty$, driving $S_l^{'} \rightarrow \infty$, which is consistent with $Z^{'} \rightarrow 1$. Thus, ER preserves a strong error signal and can correct outputs whose sign opposes the target in essentially a single update.

\subsubsection{Pseudo-inverse transport}
In ER, the transpose operations used in BP are replaced by pseudo-inverse operations, which further accelerates learning.
More precisely, the update rules $\Delta W_l = Z_{l-1}^\dagger \Delta S_l$ and $\Delta W_l = Z_{l-1}^T \Delta S_l$ can be interpreted as two limiting cases of the same optimization framework, namely ridge regression. 
Since the effect of the corrected activation $Z_{l-1}^{''}$ will be discussed later, we first focus on  $\Delta W_l = Z_{l-1}^\dagger \Delta S_l$, postponing the refinement to $\Delta W_l = Z_{l-1}^{''\dagger} \Delta S_l$.

Given an activation $Z_{l-1}$ and a target pre-activation $S_l^{'}$, ridge regression finds the weight that minimizes both the $l_2$ loss between the prediction and the target, and the Frobenius norm of the weight. Thus, the optimization problem is formualted as 
\begin{equation}
\label{eq:ridge}
    W = \arg\min_W\|S_l^{'}-Z_{l-1}W\|_2^2 + \frac{1}{2 \eta}\|W\|_2^2
\end{equation}
with the analytic solution given by $$W = \Big(Z_{l-1}^T Z_{l-1} + \frac{1}{\eta} I \Big)^{-1}Z_{l-1}^T S_l^{'}.$$
Here, $\eta$ is a regularization coefficient that controls the sparsity of $W$; smaller values of $\eta$ push $W$ closer to zero.
Consider applying ridge regression to compute $\Delta W_l = W_l^{'}-W_l$ given $Z_{l-1}$ and $\Delta S_l = S_l^{'}-S_l$, which yields $$\Delta W = \Big(Z_{l-1}^T Z_{l-1} + \frac{1}{\eta} I \Big)^{-1}Z_{l-1}^T \Delta S_l.$$
It is noteworthy that as $\eta$ decreases, the resulting $\Delta W_l$ approaches the BP solution, $$\Delta W\approx \eta Z_{l-1}^T \Delta S_l,$$
which is accurate up to the constant factor of $\eta$. Conversely, in the limit $\eta \to \infty$, we obtain $$\Delta W_l=  (Z_{l-1}^T Z_{l-1})^{-1}Z_{l-1}^T \Delta S_l=Z_{l-1}^\dagger \Delta S_l,$$
which corresponds to the linear regression solution used in the ER algorithm. This reveals a key insight: the weight updates in ER and BP arise as two opposite extremes of the same optimization framework. Moreover, the ER solution in the limit $\eta \to \infty$ allows for large weight updates $\Delta W_l$, as it is no longer significantly constrained by the regularization term on $W_l$ in Eq.~\eqref{eq:ridge}.

\subsubsection{Internal covariate shift}
Deep learning can be viewed as aligning the {\it forward signal} (input $\to$ output) with the {\it backward correction signal} (output $\to$ input) so that the network’s internal representations become consistent with the observed targets~\citep{bengio2014auto}.

In the multilayer ER, optimization starts by correcting the output pre-activation $S_L^{'} \leftarrow S_L$, and then propagating this correction backward to obtain corrected pre-activations $S_{L-1}^{'}, S_{L-2}^{'}, \cdots, S_1^{'}$.
As a result, the updated output $S_L^{'}$ is constructed to yield an ouptut activation $Z_L^{'}$ that is closer to the observation $\hat{Y}$.
Given the corrected pre-activation $S_l^{'}$, a natural next step is to form the corresponding corrected activation $Z_l^{''} = \sigma(S_l^{'})$.
This $Z_l^{''}$ represents the activation that is {\it consistent with the corrected state} at layer $l$, and it should therefore be used when updating the neighboring weights.

A key complication in multilayer networks is that updating lower-layer weights changes the inputs seen by higher layers.
If we update $W_l^{'} \leftarrow W_l$ while implicitly treating the input $Z_{l-1}$ as fixed, then the forward signal $S_l^{'} = Z_{l-1} W_l^{'}$ can become misaligned with the corrected signal $S_l^{'}$ made during the update, an effect related to ``internal covariat shift''~\citep{ioffe2015batch}.
For this reason, our update in Eq.~\eqref{eq:ERBack5} is designed to optimize $W_l$ using the {\it corrected} pair ($Z_{l-1}^{''}, S_l^{'}$), rather than the old activations.

Ideally, the updated weights $W_l^{'}$ would satisfy $S_l^{'} = Z_{l-1}^{''} W_l^{'}$. In practice, however, the pseudo-inverse step $W_l^{'} = Z_{l-1}^{''\dagger} S_l^{'}$ may not produce this relation exactly, because the pseudo-inverse is only an approximate inverse. To keep the forward signal and the weight updates well aligned despite this small mismatch, we adopt a simple and robust strategy:
We update each layer sequentially and recompute activations using the newly updated weights.
Concretely, we first update $W_1^{'} = Z_0^{\dagger} S_1^{'}$ from the given input $Z_0 = X$, and then recompute the first-layer activation as $Z_1^{''} = \sigma(Z_0 W_1^{'})$. 
This recomputation naturally incorporates any imperfection in the pseudo-inverse, unlike the direct assignment $Z_1^{''} = \sigma(S_1^{'})$.
We then proceed similarly for higher layers: after updating $W_2^{'} = Z_1^{''\dagger} S_2^{'}$, we recompute $Z_2^{''} = \sigma(Z_1^{''} W_2^{'})$, and continue layer by layer. 
In this way, each corrected activation is always formed from the most up-to-date corrected input and weights, so higher layers can smoothly adapt to the changes introduced below.

For clarity, we summarize this forward-consistent multilayer ER procedure in Algorithm~\ref{alg:cap}. For later performance comparisons, we refer to the original ER procedure based on Eqs.~\eqref{eq:ERBack1}--\eqref{eq:ERBack5} as Algorithm~0.

\begin{algorithm}
\caption{Enhanced Multilayer ER Algorithm for One Iteration}\label{alg:cap}
\begin{algorithmic}
\For{$l \gets 1$ to $L$}  \Comment{Forward process 1}
\State $S_l = Z_{l-1} W_l$ \Comment{$Z_0 = X$}
\State $Z_l = \sigma(S_l)$ \Comment{$Z_L = Y$}
\EndFor
\State $\Delta Z_L \gets \Delta Y =\hat{Y} - Y$ \Comment{Backward process}
\For{$l \gets L$ to $2$} 
\State $\Delta S_l = \dfrac{S_l}{Z_l} \odot \Delta Z_l$
\State $\Delta Z_{l-1} =\Delta S_l W_l^\dagger$ 
\EndFor
\State $\Delta S_1 = \dfrac{S_1}{Z_1} \odot \Delta Z_1$
\For{$l \gets 1$ to $L$} \Comment{Parameter update}
\State $\Delta W_l = Z_{l-1}^{\dagger} \Delta S_l$ \Comment{Note that $Z_{l-1}$ is the updated activation.}
\State $W_l \leftarrow W_l + \Delta W_l$
\State $S_l = Z_{l-1}W_l$ \Comment{Forward process 2}
\State $Z_l = \sigma(S_l)$ \Comment{Henceforth, Forward process 1 can be skipped.}
\EndFor
\end{algorithmic}
\end{algorithm}

\section{Experiments}\label{sec2}
We evaluate multilayer ER under full-batch and mini-batch learning, and compare its performance with BP.

\subsection{Full-batch}
We implemented the algorithms in PyTorch~\citep{NEURIPS2019_bdbca288}, and evaluated them on image-classification benchmarks, MNIST~\citep{lecun-mnist}.
%and CIFAR-10~\citep{krizhevsky2009learning}. 
We used fully connected neural networks with four layers, configured as follows: $784-1750-475-10$.
%for MNIST and $3072-2750-250-10$ for CIFAR-10.
Our results are not sensitive to this particular architectural choice and remain qualitatively similar across a range of network configurations.

\begin{figure}[h]
    \hfill
	\begin{center}
	\includegraphics[width=\textwidth]{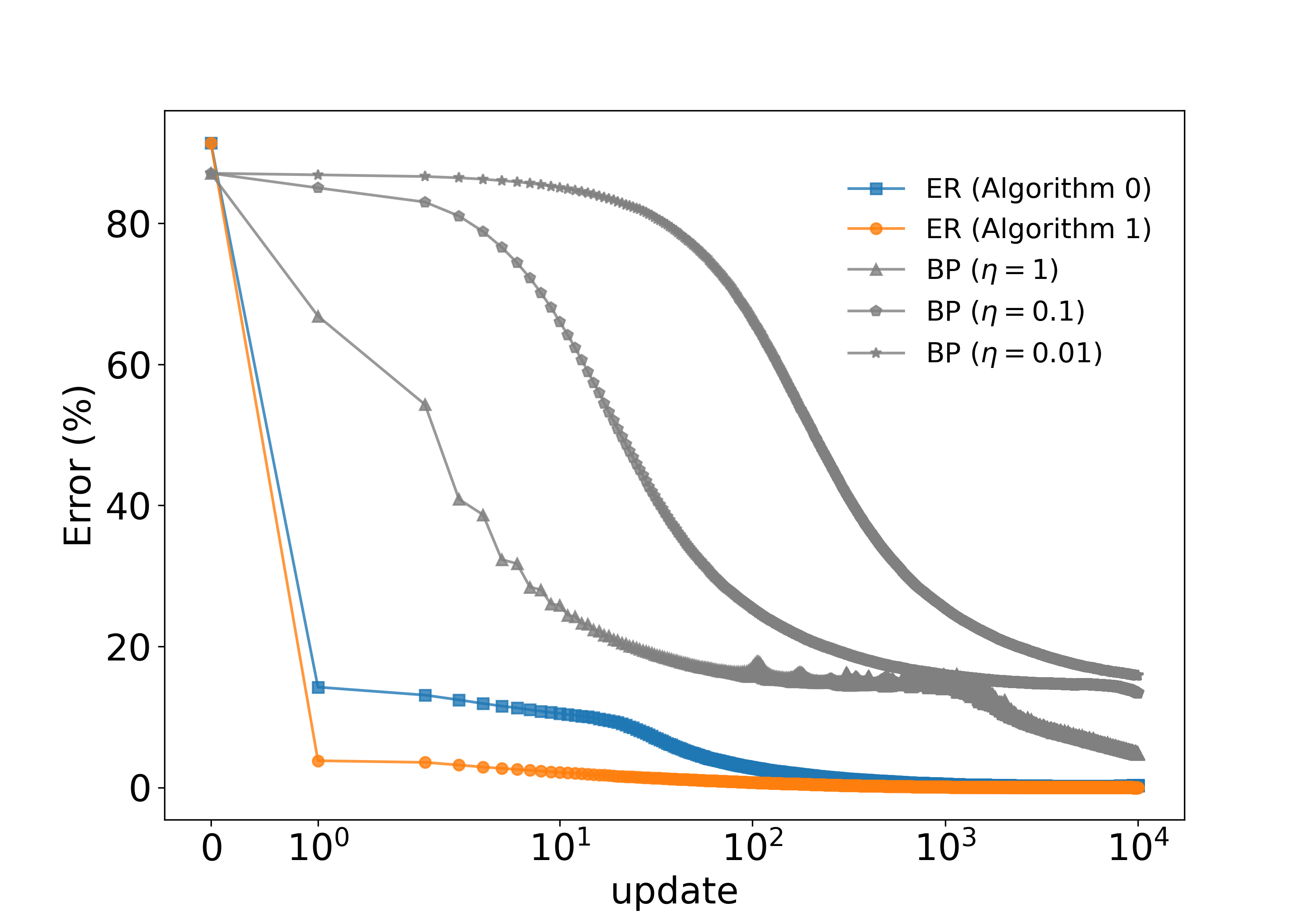}
	\end{center}
	\caption{Learning trajectories of additive and multiplicative optimization methods. Training error on the full MNIST dataset (60,000 samples) is shown as a function of the number of parameter updates for Expectation Reflection (ER) and backpropagation (BP). Two variants of ER are compared: vanilla ER (Algorithm~0; blue squares) and enhanced ER (Algorithm~1; orange circles). For BP, results are shown for different learning rates: $\eta = 1$ (gray triangles), $\eta = 0.1$ (gray octagons), and $\eta = 0.01$ (gray stars).}\label{fig:MnistFull}
\end{figure}

We first trained the network in a full-batch setting using all 60{,}000 training samples and assessed performance on the held-out 10{,}000 test samples.
Figure~\ref{fig:MnistFull} shows the evolution of the classification error rate as a function of the number of parameter updates. For reference, random guessing in a 10-class classification problem yields an expected error rate of 90\%.
Both BP and ER substantially reduce the MNIST classification error.
For BP, we explored different learning rates ($\eta = 0.01, 0.1, 1$) to examine the impact of hyperparameter on learning dynamics, whereas ER has no tunable hyperparameters.
Smaller learning rates produce smoother but slower convergence, whereas a large learning rate ($\eta=1$) leads to faster initial progress at the cost of pronounced fluctuations.
Since our primary goal is to compare the efficiency of parameter optimization between BP and ER, we focus on training error, which directly reflects the update dynamics. The corresponding test error exhibits similar trends (data not shown). The final test error reaches 3\% for ER (Algorithm~1) and 5 \% for BP ($\eta =1$).  
Notably, the learning dynamics differ markedly between the two methods. BP improves gradually, whereas ER achieves a substantial reduction in error after a single update, as anticipated. 
For a direct comparison of the parameter updates formulated in Eq.~\eqref{eq:ERBack1}-\eqref{eq:ERBack5}, we deliberately do not employ advanced gradient-based optimization techniques such as the Adam optimizer~\citep{kingma2014adam}.
We further compare the original ER (Algorithm~0) with the enhanced variant (Algorithm~1). Although Algorithm~1 performs slightly better, the improvement over Algorithm~0 remains modest.

\begin{figure}[h]
    \hfill
	\begin{center}
	\includegraphics[width=\textwidth]{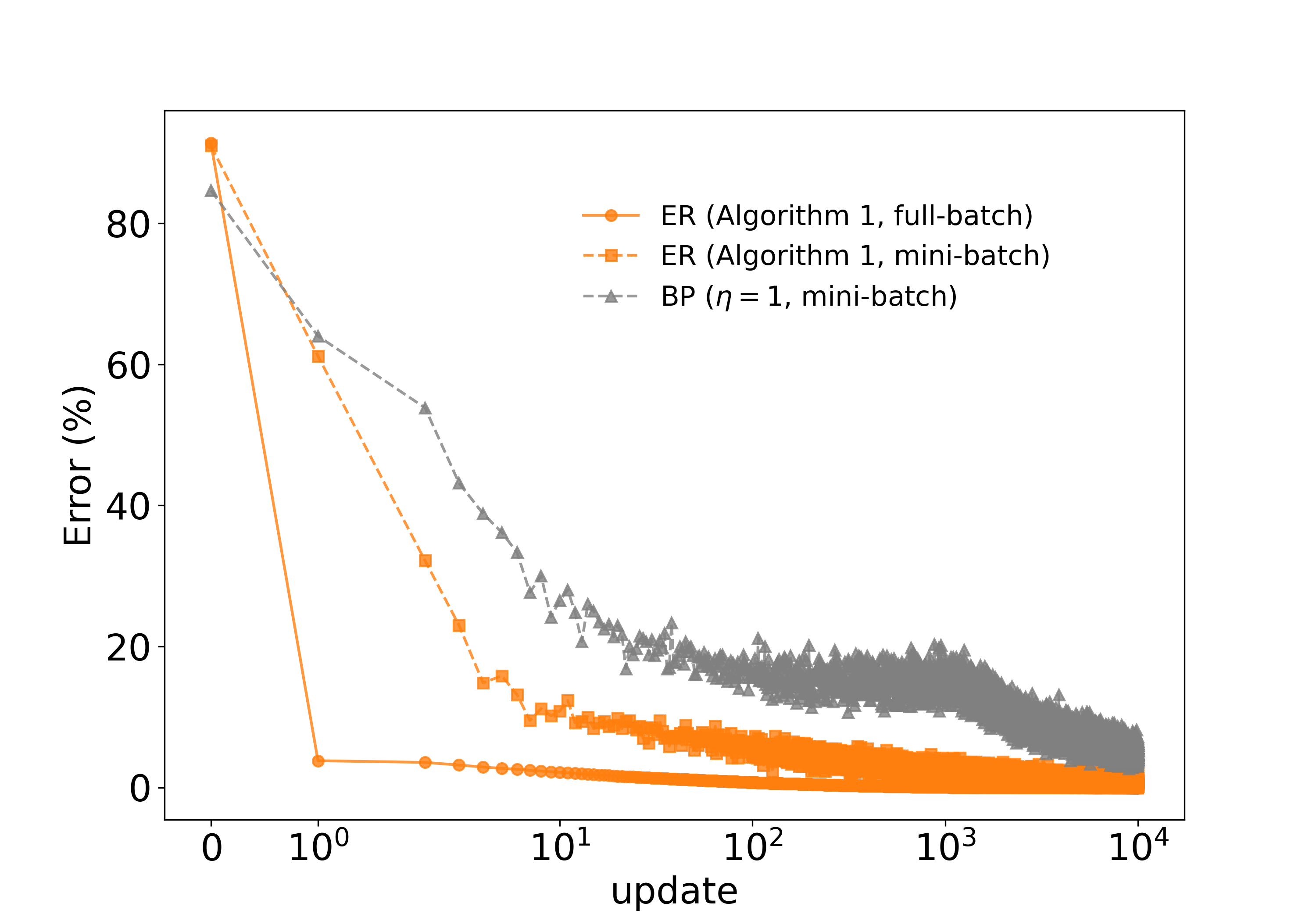}
	\end{center}
	\caption{Learning trajectories under mini-batch training. Training error on MNIST mini-batches (each containing 600 samples) is shown for Expectation Reflection (ER, Algorithm~1; orange squares with dotted lines) and backpropagation (BP with learning rate $\eta = 1$; gray triangles with dotted lines). The full-batch result of ER (Algorithm~1) from Fig.~\ref{fig:MnistFull} is included for reference (orange circles with a solid line).}\label{fig:MnistMini}
\end{figure}

\subsection{Mini-batch}
We have so far applied multilayer ER in a full-batch setting. In contrast, gradient-based methods are typically trained using randomly sampled mini-batches that change from iteration to iteration. Mini-batch learning offers two main benefits: it reduces the computational cost of processing the entire dataset at once, and the injected stochasticity can help the optimization escape poor local minima~\citep{bishop2006pattern}.

To study ER in this regime, we evaluated multilayer ER under mini-batch training. While ER is stable and highly effective in the full-batch setting, small mini-batches can introduce numerical instabilities because ER repeatedly relies on pseudo-inverse computations and can produce large one-step weight changes. Below we describe two practical modifications that stabilize mini-batch ER. Importantly, these modifications do not change the core update mechanism of ER: each iteration still computes a solve-based local estimate $W_l^{'}$ via a (regularized) pseudo-inverse, rather than taking a gradient step. The additional parameters introduced here are used only to regularize and aggregate noisy mini-batch estimates, i.e., to obtain a stable stochastic approximation to the full-batch ER update.

First, recall the pseudo-inverse $X^\dagger = (X^T X)^{-1}X^T$, which is used repeatedly in ER.
In a mini-batch setting, the batch matrix $X$ can be small, making the  covariance $X^TX$ ill-conditioned or singular and thereby causing  numerical instabilities.
To mitigate this issue, we use a ridge-regularized pseudo-inverse, $$X^\dagger = (X^T X+ \alpha I)^{-1}X^T.$$
While this modification introduces a hyperparameter ($\alpha=1$ in this study), it stabilizes the pseudo-inverse computation and improves the robustness of ER under small mini-batches.

Second, ER can update network weights substantially after a single update. This is an advantage in full-batch training, but under mini-batching the resulting batch-to-batch variability can make training unstable. To temper these abrupt changes, we aggregate successive mini-batch estimates using a weighted update,
\begin{align*}
	W_l^{(t+1)} &= (1-\eta)W_l^{(t)} + \eta W^{'}\\
	&= W_l^{(t)} +\eta (W_l^{'} - W_l^{(t)})\\
    &= W_l^{(t)} +\eta\Delta W_l^{(t)}.
\end{align*}
Here $\eta$ controls how strongly a single mini-batch estimate $W_l^{'}$ is incorporated into the running weights, and can be viewed as an averaging (trust) parameter for noisy mini-batch solves.
With these modifications, mini-batch ER (using $\eta = 0.1$ with a batch size of 600) achieves an MNIST error rate comparable to that of the full-batch setting, albeit requiring more update steps (Fig.~\ref{fig:MnistMini}).

\section{Conclusion}
\label{Conclusion}

We introduced Expectation Reflection (ER) as a ratio-based, \emph{multiplicative} learning principle and extended it to multilayer neural networks. Unlike standard backpropagation (BP), which updates parameters via additive gradient steps derived from an explicit loss function~\citep{rumelhart1986learning}, ER updates pre-activations through observation--prediction ratios and computes weights via local regression (pseudo-inverse) solves. This formulation yields a scale-adaptive update rule and, in the full-batch regime, can rapidly reconcile the forward signal (input $\to$ output) with a backward correction signal (output $\to$ input).

From a credit-assignment perspective, multilayer ER bridges local solve-based updates with methods that avoid transposed weight transport. In particular, its backward correction can be interpreted in terms of target-propagation-style inverse mappings~\citep{bengio2014auto, lee2015difference, meulemans2020theoretical}, making explicit how corrected pre-activations and recomputed activations support consistent layerwise updates. This perspective aligns with recent efforts to improve and scale target propagation~\citep{Ernoult2022} and complements broader alternatives to strict gradient transport, including feedback-alignment-style methods~\citep{lillicrap2020backpropagation}.

We further reformulated multilayer ER in BP-like notation to highlight the key substitutions that preserve locality while enabling fast learning---ratio-based pre-activation corrections, pseudo-inverse transport, and updated activations. Building on this, we proposed a forward-consistent implementation that recomputes activations after each layer update, improving consistency between updated lower-layer representations and the inputs seen by higher layers, and mitigating effects related to internal covariate shift~\citep{ioffe2015batch}.

Empirically, multilayer ER achieves competitive classification performance on MNIST with fully connected networks, exhibiting a pronounced reduction in error after the first full-batch update. Although BP with Adam~\citep{kingma2014adam} can reach similar final performance given sufficient iterations, ER attains near-peak accuracy within only a few updates, underscoring a distinct efficiency profile. We also evaluated ER in the mini-batch regime commonly used in large-scale learning~\citep{bishop2006pattern}. In this setting, smaller batches can exacerbate ill-conditioning in pseudo-inverse computations and increase update variability; nevertheless, simple regularization and conservative aggregation strategies substantially improve robustness while retaining the same underlying layerwise solve.

A key limitation of the current approach lies in the reliance on pseudo-inverse computations, which can become numerically unstable in the presence of ill-conditioned or rank-deficient intermediate representations. This issue is particularly pronounced in small-batch or when applied to more complex datasets such as  CIFAR-10~\citep{krizhevsky2009learning}. Developing more stable or regularized alternatives to the pseudo-inverse, as well as scalable approximations, remains an important direction for future work.

Overall, ER and its multilayer variants provide a complementary optimization framework that broadens the space of learning algorithms beyond purely additive, gradient-driven training, and may be particularly attractive when rapid progress with minimal tuning is a priority.

\subsection*{Funding}
This work was supported by the National Research Foundation of Korea (NRF) grant (Grant No. 2022R1A2C1006871) (J.J.), and the Intramural Research Program of the National Institutes of Health, NIDDK (V.P.).

\bibliographystyle{APA}
\bibliography{reference}

@article{pindeda1987,
	author = {Pineda, Fernando J.},
	journal = {Physical Review Letters},
	number = {19},
	pages = {2229},
	publisher = {APS},
	title = {Generalization of back-propagation to recurrent neural networks },
	volume = {59},
	year = {1987}}

@article{rao1999predictive,
	author = {Rao, Rajesh PN and Ballard, Dana H},
	journal = {Nature neuroscience},
	number = {1},
	pages = {79--87},
	publisher = {Nature Publishing Group},
	title = {Predictive coding in the visual cortex: a functional interpretation of some extra-classical receptive-field effects},
	volume = {2},
	year = {1999}}

@article{hoang2019network,
	author = {Hoang, Danh-Tai and Song, Juyong and Periwal, Vipul and Jo, Junghyo},
	doi = {10.1103/PhysRevE.99.023311},
	issue = {2},
	journal = {Phys. Rev. E},
	month = {Feb},
	numpages = {9},
	pages = {023311},
	publisher = {American Physical Society},
	title = {Network inference in stochastic systems from neurons to currencies: Improved performance at small sample size},
	url = {https://link.aps.org/doi/10.1103/PhysRevE.99.023311},
	volume = {99},
	year = {2019}}

@inproceedings{lee2015difference,
	author = {Lee, Dong-Hyun and Zhang, Saizheng and Fischer, Asja and Bengio, Yoshua},
	booktitle = {Machine Learning and Knowledge Discovery in Databases: European Conference, ECML PKDD 2015, Porto, Portugal, September 7-11, 2015, Proceedings, Part I 15},
	organization = {Springer},
	pages = {498--515},
	title = {Difference target propagation},
	year = {2015}}

@article{bengio2014auto,
	author = {Bengio, Yoshua},
	journal = {arXiv preprint arXiv:1407.7906},
	title = {How auto-encoders could provide credit assignment in deep networks via target propagation},
	year = {2014}}

@article{meulemans2020theoretical,
	author = {Meulemans, Alexander and Carzaniga, Francesco and Suykens, Johan and Sacramento, Jo{\~a}o and Grewe, Benjamin F},
	journal = {Advances in Neural Information Processing Systems},
	pages = {20024--20036},
	title = {A theoretical framework for target propagation},
	volume = {33},
	year = {2020}}

@inproceedings{ioffe2015batch,
	author = {Ioffe, Sergey and Szegedy, Christian},
	booktitle = {International conference on machine learning},
	organization = {pmlr},
	pages = {448--456},
	title = {Batch normalization: Accelerating deep network training by reducing internal covariate shift},
	year = {2015}}

@inproceedings{NEURIPS2019_bdbca288,
 author = {Paszke, Adam and Gross, Sam and Massa, Francisco and Lerer, Adam and Bradbury, James and Chanan, Gregory and Killeen, Trevor and Lin, Zeming and Gimelshein, Natalia and Antiga, Luca and Desmaison, Alban and Kopf, Andreas and Yang, Edward and DeVito, Zachary and Raison, Martin and Tejani, Alykhan and Chilamkurthy, Sasank and Steiner, Benoit and Fang, Lu and Bai, Junjie and Chintala, Soumith},
 booktitle = {Advances in Neural Information Processing Systems},
 editor = {H. Wallach and H. Larochelle and A. Beygelzimer and F. d\textquotesingle Alch\'{e}-Buc and E. Fox and R. Garnett},
 pages = {},
 publisher = {Curran Associates, Inc.},
 address   = {Red Hook, NY},
 title = {PyTorch: An Imperative Style, High-Performance Deep Learning Library},
 url = {\href{https://proceedings.neurips.cc/paper_files/paper/2019/file/bdbca288fee7f92f2bfa9f7012727740-Paper.pdf}{https://proceedings.neurips.cc/paper\_files/paper/2019/file/bdbca288fee7f92f2bfa9f7012727740-Paper.pdf}},
 volume = {32},
 year = {2019}
}

@article{lecun-mnist,
	author = {LeCun, Yann and Cortes, Corinna},
	howpublished = {http://yann.lecun.com/exdb/mnist/},
	lastchecked = {2016-01-14 14:24:11},
	title = {{MNIST} handwritten digit database},
	url = {http://yann.lecun.com/exdb/mnist/},
	username = {mhwombat},
	year = 2010}

@article{krizhevsky2009learning,
	author = {Krizhevsky, Alex and Hinton, Geoffrey and others},
	publisher = {Toronto, ON, Canada},
	title = {Learning multiple layers of features from tiny images},
	year = {2009}}

@book{bishop2006pattern,
	author = {Bishop, Christopher M and Nasrabadi, Nasser M},
	number = {4},
	publisher = {Springer},
    address = {New York},
	title = {Pattern recognition and machine learning},
	volume = {4},
	year = {2006}}

@article{rumelhart1986learning,
	author = "Rumelhart, David E and Hinton, Geoffrey E and Williams, Ronald J",
	journal = "Nature",
	number = "6088",
	pages = "533--536",
	title = "Learning representations by back-propagating errors",
	volume = "323",
	year = "1986"}

@article{lillicrap2016random,
	author = {Lillicrap, Timothy P and Cownden, Daniel and Tweed, Douglas B and Akerman, Colin J},
	journal = {Nature communications},
	number = {1},
	pages = {13276},
	publisher = {Nature Publishing Group UK London},
	title = {Random synaptic feedback weights support error backpropagation for deep learning},
	volume = {7},
	year = {2016}}

@article{kingma2014adam,
	author = {Kingma, Diederik P and Ba, Jimmy},
	journal = {arXiv preprint arXiv:1412.6980},
	title = {Adam: A method for stochastic optimization},
	year = {2014}}

@article{brown2020language,
	author = {Brown, Tom and Mann, Benjamin and Ryder, Nick and Subbiah, Melanie and Kaplan, Jared D and Dhariwal, Prafulla and Neelakantan, Arvind and Shyam, Pranav and Sastry, Girish and Askell, Amanda and others},
	journal = {Advances in neural information processing systems},
	pages = {1877--1901},
	title = {Language models are few-shot learners},
	volume = {33},
	year = {2020}}

@inproceedings{ramesh2021zero,
	author = {Ramesh, Aditya and Pavlov, Mikhail and Goh, Gabriel and Gray, Scott and Voss, Chelsea and Radford, Alec and Chen, Mark and Sutskever, Ilya},
	booktitle = {International Conference on Machine Learning},
	organization = {PMLR},
	pages = {8821--8831},
	title = {Zero-shot text-to-image generation},
	year = {2021}}

@inproceedings{rombach2022high,
	author = {Rombach, Robin and Blattmann, Andreas and Lorenz, Dominik and Esser, Patrick and Ommer, Bj{\"o}rn},
	booktitle = {Proceedings of the IEEE/CVF conference on computer vision and pattern recognition},
	pages = {10684--10695},
	title = {High-resolution image synthesis with latent diffusion models},
	year = {2022}}

@article{lillicrap2020backpropagation,
	author = {Lillicrap, Timothy P and Santoro, Adam and Marris, Luke and Akerman, Colin J and Hinton, Geoffrey},
	journal = {Nature Reviews Neuroscience},
	number = {6},
	pages = {335--346},
	publisher = {Nature Publishing Group UK London},
	title = {Backpropagation and the brain},
	volume = {21},
	year = {2020}}

@article{Crick1989,
  author  = {Crick, Francis},
  title   = {The recent excitement about neural networks},
  journal = {Nature},
  year    = {1989},
  volume  = {337},
  number  = {6203},
  pages   = {129--132},
  doi     = {10.1038/337129a0}
}

@article{Whittington2019,
  author  = {Whittington, James C. R. and Bogacz, Rafal},
  title   = {Theories of Error Back-Propagation in the Brain},
  journal = {Trends in Cognitive Sciences},
  year    = {2019},
  volume  = {23},
  number  = {3},
  pages   = {235--250},
  doi     = {10.1016/j.tics.2018.12.005}
}

@book{Hebb1949,
  author    = {Hebb, Donald O.},
  title     = {The Organization of Behavior: A Neuropsychological Theory},
  publisher = {Wiley},
  address   = {New York},
  year      = {1949}
}

@article{Caporale2008,
  author  = {Caporale, Natalie and Dan, Yang},
  title   = {Spike timing-dependent plasticity: a {Hebbian} learning rule},
  journal = {Annual Review of Neuroscience},
  year    = {2008},
  volume  = {31},
  pages   = {25--46},
  doi     = {10.1146/annurev.neuro.31.060407.125639}
}

@article{Fremaux2016,
  author  = {Fr{\'e}maux, Nicolas and Gerstner, Wulfram},
  title   = {Neuromodulated Spike-Timing-Dependent Plasticity, and Theory of Three-Factor Learning Rules},
  journal = {Frontiers in Neural Circuits},
  year    = {2016},
  volume  = {9},
  pages   = {85},
  doi     = {10.3389/fncir.2015.00085}
}

@article{Amari1998,
  author  = {Amari, Shun-ichi},
  title   = {Natural Gradient Works Efficiently in Learning},
  journal = {Neural Computation},
  year    = {1998},
  volume  = {10},
  number  = {2},
  pages   = {251--276},
  doi     = {10.1162/089976698300017746}
}

@book{NemirovskiYudin1983,
  author    = {Nemirovski, Arkadi and Yudin, David B.},
  title     = {Problem Complexity and Method Efficiency in Optimization},
  publisher = {Wiley},
  address   = {New York},
  year      = {1983}
}

@article{BeckTeboulle2003,
  author  = {Beck, Amir and Teboulle, Marc},
  title   = {Mirror Descent and Nonlinear Projected Subgradient Methods for Convex Optimization},
  journal = {Operations Research Letters},
  year    = {2003},
  volume  = {31},
  number  = {3},
  pages   = {167--175},
  doi     = {10.1016/S0167-6377(02)00231-6}
}

@article{Soh2023MirrorHopfield,
  author  = {Soh, H. and others},
  title   = {Mirror Descent of Hopfield Model},
  journal = {Neural Computation},
  year    = {2023},
  volume  = {35},
  number  = {9},
  pages   = {1529--},
}

@article{KivinenWarmuth1997,
title = {Exponentiated Gradient versus Gradient Descent for Linear Predictors},
journal = {Information and Computation},
volume = {132},
number = {1},
pages = {1-63},
year = {1997},
issn = {0890-5401},
doi = {https://doi.org/10.1006/inco.1996.2612},
url = {https://www.sciencedirect.com/science/article/pii/S0890540196926127},
author = {Jyrki Kivinen and Manfred K. Warmuth},
}

@article{LeeSeung2001,
  author  = {Lee, Daniel D. and Seung, H. Sebastian},
  title   = {Algorithms for Non-negative Matrix Factorization},
  journal = {Advances in Neural Information Processing Systems},
  year    = {2001},
  volume  = {13},
  pages   = {556--562}
}

@article{DarrochRatcliff1972,
  author  = {Darroch, J. N. and Ratcliff, D.},
  title   = {Generalized Iterative Scaling for Log-Linear Models},
  journal = {The Annals of Mathematical Statistics},
  year    = {1972},
  volume  = {43},
  number  = {5},
  pages   = {1470--1480},
  doi     = {10.1214/aoms/1177692379}
}

@article{Han2021,
title = {Mechanistic gene networks inferred from single-cell data with an outlier-insensitive method},
journal = {Mathematical Biosciences},
volume = {342},
pages = {108722},
year = {2021},
issn = {0025-5564},
doi = {https://doi.org/10.1016/j.mbs.2021.108722},
url = {https://www.sciencedirect.com/science/article/pii/S0025556421001322},
author = {Jungmin Han and Sudheesha Perera and Zeba Wunderlich and Vipul Periwal},
}

@article{Penrose1955,
  author  = {Penrose, Roger},
  title   = {A Generalized Inverse for Matrices},
  journal = {Proceedings of the Cambridge Philosophical Society},
  year    = {1955},
  volume  = {51},
  number  = {3},
  pages   = {406--413},
  doi     = {10.1017/S0305004100030401}
}

@inproceedings{Ernoult2022,
  author    = {Ernoult, Maxence and Normandin, Fabrice and Moudgil, Abhinav and Spinney, Sean and Belilovsky, Eugene and Rish, Irina and Richards, Blake A. and Bengio, Yoshua},
  title     = {Towards Scaling Difference Target Propagation by Learning Backprop Targets},
  booktitle = {Proceedings of the 39th International Conference on Machine Learning},
  series    = {Proceedings of Machine Learning Research},
  year      = {2022},
  publisher = {PMLR},
  url       = {https://proceedings.mlr.press/v162/ernoult22a.html}
}

\end{document}